\documentclass[fleqn,10pt]{wlscirep}
\usepackage[utf8]{inputenc}
\usepackage[T1]{fontenc}
\usepackage{xurl}

\usepackage{subcaption}

\title{Comparison of marker-less 2D image-based methods for infant pose estimation}

\author[1,2,*]{Lennart Jahn}
\author[1]{Sarah Fl\"ugge}
\author[3,4]{Dajie Zhang}
\author[3]{Luise Poustka}
\author[5,6,7]{Sven B\"olte}
\author[2]{Florentin W\"org\"otter}
\author[1,3,4,5]{Peter B Marschik}
\author[1,2,3]{Tomas Kulvicius}

\affil[1]{Child and Adolescent Psychiatry and Psychotherapy, University Medical Center Göttingen; German Center for Child and Adolescent Health (DZKJ), Leibniz ScienceCampus Göttingen, Von-Siebold-Str. 5, Göttingen, Germany}
\affil[2]{University of Göttingen, III Institute of Physics - Biophysics, Göttingen, Germany}
\affil[3]{Department of Child and Adolescent Psychiatry, University Hospital Heidelberg, Ruprecht-Karls University of Heidelberg, Heidelberg, Germany}
\affil[4]{iDN – interdisciplinary Developmental Neuroscience, Division of Phoniatrics, Medical University of Graz, Graz, Austria}
\affil[5]{Center of Neurodevelopmental Disorders (KIND), Department of Women’s and Children’s Health, Center for Psychiatry Research, Karolinska Institutet \& Region Stockholm, Stockholm, Sweden}
\affil[6]{Child and Adolescent Psychiatry, Stockholm Health Care Services, Region Stockholm, Stockholm, Sweden}
\affil[7]{Curtin Autism Research Group, Curtin School of Allied Health, Curtin University, Perth, Australia}

\affil[*]{lennart.jahn@phys.uni-goettingen.de}


\begin{abstract}
In this study we compare the performance of available generic- and specialized infant-pose estimators for a video-based automated general movement assessment (GMA), and the choice of viewing angle for optimal recordings, i.e., conventional diagonal view used in GMA vs. top-down view. We used 4500 annotated video-frames from 75 recordings of infant spontaneous motor functions from 4 to 16 weeks. To determine which pose estimation method and camera angle yield the best pose estimation accuracy on infants in a GMA related setting, the distance to human annotations and the percentage of correct key-points (PCK) were computed and compared. The results show that the best performing generic model trained on adults, ViTPose, also performs best on infants. We see no improvement from using specific infant-pose estimators over the generic pose estimators on our infant dataset. However, when retraining a generic model on our data, there is a significant improvement in pose estimation accuracy. This indicates limited generalization capabilities of infant-pose estimators to other infant datasets, meaning that one should be careful when choosing infant pose estimators and using them on infant datasets which they were not trained on. The pose estimation accuracy obtained from the top-down view is significantly better than that obtained from the diagonal view (the standard view for GMA). This suggests that a top-down view should be included in recording setups for automated GMA research.
\end{abstract}
\begin{document}

\flushbottom
\maketitle

\thispagestyle{empty}

\section{Introduction}
\subsection{Background}
When classic biomarker approaches fail to detect developmental conditions early or predict neurodevelopmental outcomes following pre- or perinatal brain lesions or complications during pregnancy, assessments of overt neurofunctions come into play \cite{Marschik_interdisciplinaryquestbehavioral_2016a,Einspieler_GeneralMovementOptimality_2024,Novak_EarlyAccurateDiagnosis_2017}. An early detection of biological markers  of adverse outcomes or early diagnosis of developmental conditions facilitates early intervention, when brain plasticity is at its peak, aiming to achieve best possible long-term outcomes \cite{Kirton_Perinatalstrokemapping_2021}. The Prechtl general movement assessment (GMA) \cite{prechtl1997early} relies on human visual gestalt perception to delineate the spontaneous motor repertoire which is continuously present from fetal life to about 5 months post-term age \cite{einspieler2008human,Einspieler_Fetalmovementsorigin_2021}. The fetus and the newborn move without sensory stimulation in age-specific patterns, so called general movements (GMs). These movements involve the whole body, appear as writhing movements from about 6-9 weeks post-term age and change their character to so-called fidgety movements later on. Neurobiology has coined the term ‘central pattern generator’ (CPG) for the underlying neural circuitry of endogenously generated activity that is not triggered by sensory stimulation but can be modulated by the periphery. General movements that present with a variable sequence of arm, leg, neck, and trunk movements have been studied through human and augmented Gestalt perception approaches \cite{Irshad_AIApproachesPrechtl_2020,silva2021future,Marschik-Zhang_BuildingBlocksDeep_2023,Kulvicius_Deeplearningempowered_2025}. Over the last decades, GMA widened its scope and applicability to serve as a general estimate of the integrity of the developing nervous system. It became a tool to look beyond the mere distinction of typical development vs. high risk for cerebral palsy, e.g., to determine the impact of viral infections on postnatal neurodevelopment or describe the early phases of neurodevelopmental conditions such as autism \cite{Einspieler_Highlightingfirstmonths_2014,einspieler2019association}. This broad applicability was the reason we used the multi-camera recording technique for a population based study which included the standard GMA perspective as a use case  \cite{einspieler2004prechtl,einspieler2019association,Marschik-Zhang_BuildingBlocksDeep_2023}. GMA is a gestalt-based observational tool to classify spontaneous infant motor functions in the first months of life \cite{prechtl1997early}. Infants up to 20 weeks of post-term age are positioned in supine position in a cot and their spontaneous movements, i.e. unstimulated, are video-recorded. These movements can be analyzed and classified into physiological or atypical movement patterns \cite{prechtl1997early,einspieler2008human,Einspieler_GeneralMovementOptimality_2024,Novak_EarlyAccurateDiagnosis_2017}. 
Given its high sensitivity and specificity GMA has become one tool of choice for the early detection of cerebral palsy in the early postnatal period \cite{Novak_EarlyAccurateDiagnosis_2017}. 

\subsection{State of the Art}
\label{sec:sota}
Recently, efforts have been intensified to automate GMA using computers (for recent reviews see \cite{irshad2020ai,silva2021future,raghuram2021automated,leo2022video,peng2023continuous,reinhart2024artificial}).
In early studies, spontaneous movements in the first months of life were measured with sensors directly attached to the infant \cite{Karch_Kinematicassessmentstereotypy_2012,Philippi_Computerbasedanalysisgeneral_2014}. However, the required procedures and the presence of the sensors could potentially change motion \cite{silva2021future}, thus, efforts to augment or replace wearable sensors with non-invasive computer vision based methods have become increasingly popular.
Early camera based works used optical flow methods \cite{adde2009using,adde2010early,Orlandi_DetectionAtypicalTypical_2018,Tsuji_MarkerlessMeasurementEvaluation_2020} for motion detection. 
With the advent of neural networks for pose estimation \cite{Mathis_DeepLabCutmarkerlesspose_2018,Cao_OpenPoseRealtimeMultiPerson_2021,Groos_EfficientPoseScalablesingleperson_2021,Xu_ViTPoseSimpleVision_2022a} current studies have switched to pose estimation and skeleton keypoints for visual-based movement analysis and classification \cite{McCay_AbnormalInfantMovements_2020,Reich_NovelAIdriven_2021a,nguyen2021spatio,Groos_DevelopmentValidationDeep_2022,Moro_markerlesspipelineanalyze_2022,Marschik_Openvideodata_2023,gao2023automating,Morais_RobustInterpretableGeneral_2023,Morais_FinegrainedFidgetyMovement_2024}.
At first, most automated GMA studies \cite{McCay_AbnormalInfantMovements_2020,Reich_NovelAIdriven_2021a,nguyen2021spatio,Marschik_Openvideodata_2023} used OpenPose \cite{wei_convolutional_2016, cao_realtime_2017, Cao_OpenPoseRealtimeMultiPerson_2021}. 
More recent approaches changed to newer pose estimators like EfficientPose \cite{Groos_DevelopmentValidationDeep_2022}, the adaptable pose estimation framework DeepLabCut \cite{Moro_markerlesspipelineanalyze_2022}, or a derivative version of OpenPose fine tuned on infant datasets \cite{Morais_RobustInterpretableGeneral_2023,Morais_FinegrainedFidgetyMovement_2024}.

Recent findings revealed ViTPose to be the best performing framework for human pose estimation using the COCO dataset  \cite{Xu_ViTPoseSimpleVision_2022a}.
However, all current pose estimators with large underlying datasets are trained on adults and then used for infant pose estimation without modification. One of the reasons is that there are no large public infant datasets available.
The field of infant pose estimation for clinical applications in general faces a complex data sharing issue which presents a true obstacle for all video-based clinical methods \cite{Marschik_Openvideodata_2023}. Infant specific models \cite{Moccia_Preterminfantspose_2020,Huang_InvariantRepresentationLearning_2021,Cao_AggPoseDeepAggregation_2022,Soualmi_3Dposeestimation_2023,Yin_selfsupervisedspatiotemporalattention_2024} are consequently mostly trained and tested on silo-datasets (patient data) and results have not yet been compared on a common test-set.
There are comparisons between related model architectures when selecting the best model \cite{Cao_AggPoseDeepAggregation_2022,Soualmi_3Dposeestimation_2023}, but they are all trained on the same data.
There is also a comparison of different existing models on infant data\cite{Gama_Automaticinfant2D_2024}, however, there, no infant data was used in training.
Thus, it is unknown if the generalization capabilities of specialized infant pose estimators are sufficient for use on datasets the particular models were not trained on.

Additionally, GMA, in its clinical application, is usually done with single 2D-RGB cameras in a diagonal view \cite{einspieler2004prechtl,marschik2017novel}. This viewing angle is uncommon for pictures of adults on which generic pose estimators are trained, which might also have effects on the pose estimators' performance on infant datasets.

\subsection{Contribution}

In this work, we compare various infant pose estimators (generic and infant specific) to evaluate the performance increase using state of the art methods. We also analyze the generalization capabilities of infant specific pose estimators on datasets different from training, and quantify the effect of different viewing angles on pose estimation accuracy. This is important, as selection of the best pose estimator will directly influence the quality of any task done on the extracted poses.
\newline

Specifically, we address the following four main research questions:
\begin{itemize}
    \item Which generic pose estimator is best suited for infant pose estimation?
    
    To test the generalization capabilities of generic pose estimation models, and to later compare those to the infant specific models, we first analyzed four different pose estimation models, OpenPose \cite{wei_convolutional_2016, cao_realtime_2017, Cao_OpenPoseRealtimeMultiPerson_2021}, MediaPipe \cite{google_mediapipe_nodate,Bazarevsky_BlazePoseOndeviceRealtime_2020}, HRNet \cite{wang_deep_2019, sun_deep_2019} and ViTPose \cite{Xu_ViTPoseSimpleVision_2022a}.
    Specific reasons for the inclusion of all models are given in section \ref{sec:methods_pose_frameworks}. We find that ViTPose performs best.

    \item Do other infant pose estimators generalize well enough such that they achieve better accuracy on our dataset, or is specific retraining necessary?

    We evaluated the infant-specific pose estimators AggPose \cite{Cao_AggPoseDeepAggregation_2022} and AGMA-HRNet48 \cite{Soualmi_3Dposeestimation_2023} on our dataset. In a second step, we retrained the best performing generic model ViTPose \cite{Xu_ViTPoseSimpleVision_2022a} on our dataset and then compared the retrained model to the two specific infant pose estimators.
    As detailed in \ref{sec:sota}, a comparison like this has not yet been done for infants using data from a source different from the one used in training.
    We find that retraining, as predicted, increases accuracy, but also that the infant-specific models do not generalize well and only perform at or below the level of the best generic model.
     
    \item Is there a difference in pose estimation accuracy between viewing angles (classic diagonal GMA-perspective vs. top-down)?
    
    We studied the influence of the viewing angle on the pose estimation methods by splitting the analysis by viewing angles. Since the data used in this study comes from a recording setup with multiple angles (see Figure \ref{fig:setup_overview}a) for later 3D-reconstruction, we can directly compare the different viewing angles that were commonly used in other works, a diagonal view and a top view (see Figure \ref{fig:setup_overview}b,c), and quantify their influence on pose estimation accuracy. We find better performance for the top view as compared to the diagonal (GMA preferred) view.
    
    \item Is having a dedicated estimator per view angle better than one trained on multiple views?
    
    We specifically retrained ViTPose on single view angles, to test if this further increases the accuracy, or at least leads to equal performance on top and diagonal views (for single model, the diagonal view was worse in any situation). We find that angle specific ViTPose models lead to no improvement over a unified model.
    
\end{itemize}

Our findings provide useful guidelines for the optimal selection of camera angles and pose estimation models for future projects on automated infant movement classification (such as GMA).

\section{Materials and Methods}

\subsection{Dataset}

\begin{figure*}[ht]
    \centering
    \includegraphics[width=\textwidth]{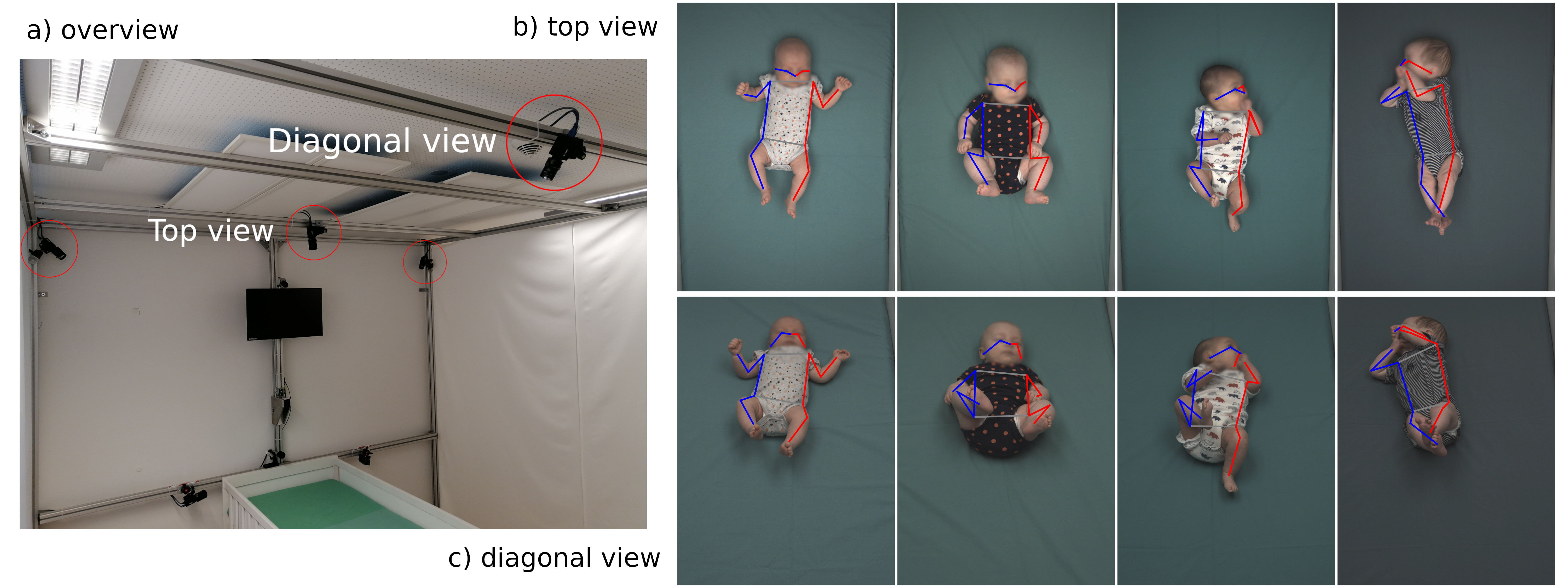}
    \caption{Overview of the recording setup a) and its output b) and c). The cameras recording the infants are circled in red. For this study, only the two cameras labeled diagonal / top view were used. Panels b) and c) show example frames for infants of different age and pose complexity from the two different views. The extracted pose keypoints are displayed as skeletons over the image. Note that neither the human annotators nor any of the pose estimators could reliably determine the position of the fully covered ear in the rightmost example.}
    \label{fig:setup_overview}
\end{figure*}

We built a dataset of 4500 frames with COCO \cite{10.1007/978-3-319-10602-1_48} style labeling (keypoints: nose, eyes, ears, shoulders, elbows, wrists, hips, knees and ankles).
The recordings for this analysis were performed utilising a multi-angle marker-less motion tracking setup, specifically designed to record young infants and include 75 recordings of 31 participants from 28 $\pm$ 2 days to 112 $\pm$ 2 days of gestational age.
For this study, we chose the two most common camera angles and relevant view points in the clinical use of GMA: straight down from right above the bed (top-down view) and a view as for a human standing at the foot end of the bed (diagonal view; standard clinical GMA view; \cite{einspieler2004prechtl}).
Figure \ref{fig:setup_overview} shows an overview of the camera setup and its output.

\subsubsection{Participants}

Participants included in this study were prospectively recruited from 2021 to 2023 in Göttingen, Germany, and its close surroundings. The umbrella project aims to investigate cross-domain ontogenetic development in early infancy. To embrace the variability of the targeted dimension, i.e., spontaneous motor functions, 31 participants (17 female) were included. The gestational age at birth of the sample ranged from 34 to 42 weeks. At the time of data analysis, no participant was diagnosed with a neurological or neuromotor deficit, nor any neurodevelopmental impairment. All parents of the participating infants provided written informed consent to study participation and publication of depersonalized data and results. The study was approved by the Institutional Review Board (Ethics Commission) of the University Medical Center Göttingen (19/2019) and performed in accordance with the relevant guidelines and regulations.

\subsubsection{Infant movements recordings}

Standard laboratory recordings of infant movements at three timepoints (T1 – T3) were included in the current study. These were extracted from data of the umbrella project, assessments, among others, infants’ spontaneous movements in a standard laboratory-setting from the 4th to the 18th week of post-term age (PTA; corrected age; from here on, ages refer to the post-term age if not otherwise specified). For the current study, available laboratory recordings (n = 75) of the participants for the following timepoints are analyzed: T1: 28 $\pm$ 2 days, T2: 84 $\pm$ 2 days, and T3: 112 $\pm$ 2 days of PTA. As known, infants at 4 weeks (corresponding to T1) present a different spontaneous movement pattern than at 12 and 16 weeks (corresponding to T2 and T3; \cite{einspieler2016fidgety, einspieler2021fetal}). With standard recordings from T1 to T3, we intended to cover the distinct age-specific spontaneous movement repertoires of the young infants \cite{einspieler2008human}.

\subsubsection{Recording Setup}
\label{sup:setup}
In total, four cameras were mounted to record infants lying in a cot: one from the top down, one diagonally from the foot end and two from the sides.
This provides robustness against occlusions and 3D-reconstruction capabilities.
The angles were fixed and did not change across the entirety of the dataset.
The recordings were done at $1600 \times 1200$ pixels with a frame rate of 60\,Hz.
All cameras had global shutter and were triggered synchronously for optimal 3D-triangulation capabilities.

For this study, we did not yet do 3D-triangulation, but first determined the optimal pose estimation method on the 2D images alone.
The setup also records two side views that are not used in standard GMA settings and were not used for computer vision in this study as well.

\subsubsection{Image selection and annotation}
\label{sup:frames}

The 75 recordings were split into two parts of 50 and 25 which were annotated by one research assistant each, using \textbf{JARVIS Annotation Tool 1.2} \cite{huser_jarvis_2022}.
Additionally, 10\% of the dataset, regardless of the split, were double-annotated for human error and individual joint labeling difficulty estimation.

To select the video frames for the dataset from the available recordings, we first selected time ranges where the infants were in a behavioural state suitable for GMA.
From those, we extracted 30 frame-sets each, by $k$-means-clustering (with $k = 30$) on sub-sampled versions of the videos.
This yielded good coverage of the different body poses displayed by the infants (see Figure \ref{fig:setup_overview} panels b) and c)).
In summary, we therefore had 30 frame-sets each obtained from 75 recordings, from which we took two different perspectives.
In total, this corresponds to $30\cdot 75\cdot 2 = 4500$ frames.

It is not always possible to see all of the keypoints in an image.
While there are some instances of eyes, shoulders or ankles being completely occluded, most of the occluded keypoints result from a physiological age-specific ATNR pattern; asymmetric tonic neck reflex. 
Across the whole dataset, 11.2\% of ears could not be seen, because the head was turned on its side.
Only counting infants of age smaller than 42 days, the percentage of missing ears was 17.2\%, compared to 7.5\% for older infants.

Some of the pose estimation frameworks we compared support more or less keypoints than COCO, but here we only evaluate the COCO-style keypoints.
The annotation software is able to use the 3D calibration information and reproject annotated points within the frame sets. So, although we only labeled the top and diagonal views, we supplied the side views to the annotators to deal better with occlusions through re-projection of annotations from the side views (if needed).

\subsection{Pose estimation frameworks}
\label{sec:methods_pose_frameworks}

\subsubsection{Generic pose estimation}
\label{sec:methods_generic_pose}

For comparison, we selected four different generic human pose estimation frameworks.
\paragraph{OpenPose \textnormal{\cite{wei_convolutional_2016, cao_realtime_2017, Cao_OpenPoseRealtimeMultiPerson_2021}}}
was one of the first pose estimation frameworks used for extraction of movements for analysis and classification of infant motor functions \cite{McCay_AbnormalInfantMovements_2020, Chambers_ComputerVisionAutomatically_2020, Reich_NovelAIdriven_2021a}.
We included this framework as a baseline method to evaluate the performance gain of state-of-the-art models.
We used the standard body 25 model available from \url{https://github.com/CMU-Perceptual-Computing-Lab/openpose}.
\paragraph{MediaPipe pose \textnormal{\cite{google_mediapipe_nodate,Bazarevsky_BlazePoseOndeviceRealtime_2020}}}
This framework is not primarily intended for accurate pose estimation, but rather for fast inference on mobile devices.
It was interesting for our comparison as Google advertises it as suitable for Yoga and fitness applications and included 25000 images of fitness exercises, so we expected it to deal well with different viewing angles and poses. We used version 0.9.2.1 as available through \texttt{pip}.
\paragraph{HRNet \textnormal{\cite{wang_deep_2019, sun_deep_2019}}}
The HR in HRNet stands for high resolution. The particular version used (HRNet-w48 available through MMPose \cite{openmmlab_mmpose_2020}, \url{https://mmpose.readthedocs.io/en/latest/model_zoo/body_2d_keypoint.html#cid-hrnet-on-coco}) has an input size of $512 \times 512$ pixels.
Because our videos are of sufficient resolution and adequate quality with low compression artefacts, this could benefit the pose estimation accuracy.
\paragraph{ViTPose \textnormal{\cite{Xu_ViTPoseSimpleVision_2022a}}}
This model is the current best performing framework for human pose estimation on the COCO test-dev dataset and uses a Vision Transformer architecture instead of the conventional CNN.
We used the version available through mmpose \cite{openmmlab_mmpose_2020} of the ``huge'' variety (\url{https://mmpose.readthedocs.io/en/latest/model_zoo/body_2d_keypoint.html#topdown-heatmap-vitpose-on-coco}).

\subsubsection{Infant pose estimation}

We also selected two pose estimation frameworks which were trained on infant images and tested them against generic frameworks. In addition, we also included a ViTPose model which was retrained on our infant dataset (see Section \ref{sec:methods_retraining}).

\paragraph{AggPose \textnormal{\cite{Cao_AggPoseDeepAggregation_2022}}}
The AggPose model is based on the Transformer architecture, like ViTPose, and is specifically trained for infant pose detection on a proprietary dataset.
In general, the images in the dataset also stem from recordings used for GMA, with infants in supine position, like in our dataset.
We used the version published by the authors on github (\url{https://github.com/PediaMedAI/AggPose}).

\paragraph{AGMA-HRNet48 \textnormal{\cite{Soualmi_3Dposeestimation_2023}}}
Soualmi et al. also retrained infant pose estimation networks on a proprietary dataset of GMA recordings.
Compared to our dataset, the images contain clinical equipment  (e.g., tubes, electrodes, cables), leading to more visual clutter in the scenes.
The authors retrained multiple networks, of which we chose the HRNet48, because it is the same architecture as the one we used for the comparison of the generic pose estimation frameworks.
We used the version published by the authors (\url{https://drive.google.com/drive/folders/1SEuTqrNdz6ubRGwMaUazil0BVOqf2cYw?usp=sharing}.)

\subsubsection{Retraining}
\label{sec:methods_retraining}

We retrained the best performing framework ViTPose (see Section \ref{sec:generic_pose}) on our labeled dataset using 5-fold cross-validation. We manually split the dataset such that all images of each infant were contained in one fold only.
This ensures the networks do not get better test results by overfitting on specific infants, which are correlated across different recording dates.
Each of the five splits therefore contained 900 frames, made up from 450 different situations recorded from two different angles.
Because of the different numbers of recordings done for each infant, four splits contained images from six different infants and one split contained images from seven different infants.
For training, we randomly split 10\% of the training set off as validation set, yielding a final training set size of 3240 frames in each fold.
For each fold, we then retrained the model using the default training parameters for ViTPose from the MMPose framework \cite{openmmlab_mmpose_2020}.
Our graphics card did not have enough memory to retrain ViTPose-huge, so we retrained the smaller version ViTPose-large instead.
We used the pretrained ViTPose model weights for initialization and employed a validation stop on PCK.
We then kept the model with the highest PCK on the validation set for evaluation on the test set.
The fold with the longest retraining took only 49 epochs until validation stop.
Compared to the training of 210 epochs on the substantially larger COCO dataset that went into the base model, this is a relatively inexpensive retraining.
All five models are published on Zenodo (\url{https://doi.org/10.5281/zenodo.14833182}).

The results presented for ''Retrained ViTPose'' are the combined test set estimations of the five individual models, which in total yield the full dataset.

\subsection{Quantification metrics}
\label{sec:methods_quantification}

We used two main evaluation metrics to assess the performance of the pose estimation methods: the difference between predicted keypoint position and the human annotation, $d_a$, in pixels, and the percentage of correct keypoints (PCK).

Our dataset is not biased to any specific side (left or right) of the infants.
Moreover, labeling left and right sides of hands, knees, etc., is not inherently different.
Therefore, we merged the evaluation results for keypoints of the left and right side variety into one.

\subsubsection{Pose estimation error}
The difference between model prediction and the human annotation ($d_a$) is the Euclidean distance between the predicted and human label in pixels:
\begin{align}
    d_a = \sqrt{(p_x - h_x)^2 + (p_y - h_y)^2},
\end{align}
where $p_{x/y}$ are the network predictions and $h_{x/y}$ the human labels for a keypoint.

\subsubsection{Percentage of Correct Keypoints (PCK)}
For PCK \cite{Yang_ArticulatedHumanDetection_2013}, all keypoint predictions are classified as either correct or incorrect, based on their distance to the label (as given above).
The PCK score is then defined as the fraction of correct keypoints out of all keypoints.
The distance threshold to be considered correct is defined as a percentage of the apparent torso length of the infant in the frame, measured as the distance between the left shoulder and hip.
In this work we considered the PCK at 5\%, 7.5\% and 10\% of the torso size, denoted by PCK@0.05, PCK@0.75 and PCK@0.1, respectively.

\subsubsection{Relation to real-world distance}
It is not possible to transform any of the quantification measures directly into real-world distances without using 3D triangulation, which was not done in this work, or assuming the points are on the same plane, which is untrue because the infants frequently lift their extremities.
For better intuition, we can still provide an upper limit estimation (i.e., the real world distance can not be greater than this) to relate our evaluation metrics to the real-world distance in our recording setup.
In our setting, the size of one pixel on the plane of the bed, \textit{always} corresponds to a distance $\leq 0.8$\,mm (in any camera view).
Therefore, 80\% of the pixel difference is an upper bound to the real-world distance in mm.
Considering the mean torso length of 306 pixels, PCK@0.1 yields the percentage of points that were detected within approximately 2.5\,cm of the ground truth.

Although both metrics cannot be directly used for real-world-distances, they are fully appropriate for relative comparisons between the pose estimation frameworks.

\subsubsection{Statistical tests}

When comparing the results with respect to the difference $d_a$, we tested the statistical significance of model differences with a paired sample t-test on all keypoints in the dataset (or sub-sets of keypoints if indicated). 
For PCK, where each point is either correct or incorrect, we used a Pearson's chi-squared test on the frequencies of the outcomes (correct detection or not) to test if the samples for each pose estimation model could come from the same distribution.
%The error bars in all figures depict the 95\% confidence intervals on the mean.

\section{Results}
\label{sec:results}

\subsection{Variability in human labeling}
\label{sec:humanvariation}

\begin{figure}[ht]
        \centering
        \includegraphics[width = 0.6\linewidth]{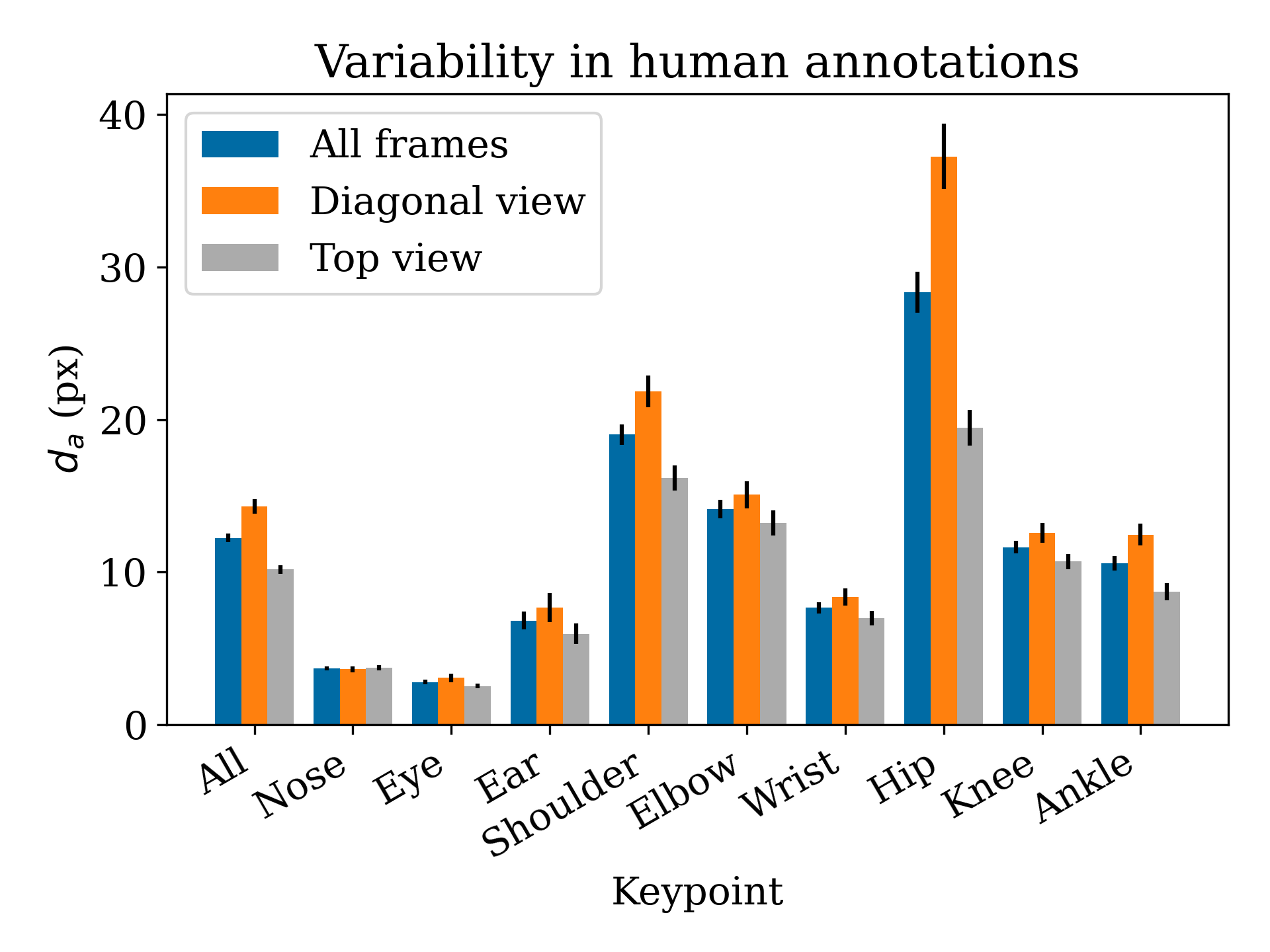}
        \caption{Difference between two annotators, additionally split by viewing angle. Error bars represent confidence intervals of mean (95\%).}
        \label{fig:humanerror}    
\end{figure}

Figure \ref{fig:humanerror} shows the mean pixel difference between the two annotators for each keypoint.
The results show that the keypoints that were hardest to label are the hips and shoulders.

When comparing the two viewing angles, there is a difference between the diagonal and top views.
The error for the diagonal view is consistently higher than for the top view.
Except for the nose keypoint, the confidence intervals of the means never overlap.
For the hip keypoint, which has the highest error overall, the relative difference between top and diagonal view is also highest.

\subsection{Generic pose estimators}
\label{sec:generic_pose}

\begin{figure*}[ht]
    \begin{subfigure}[t]{0.49\linewidth}
        \includegraphics[width = \linewidth]{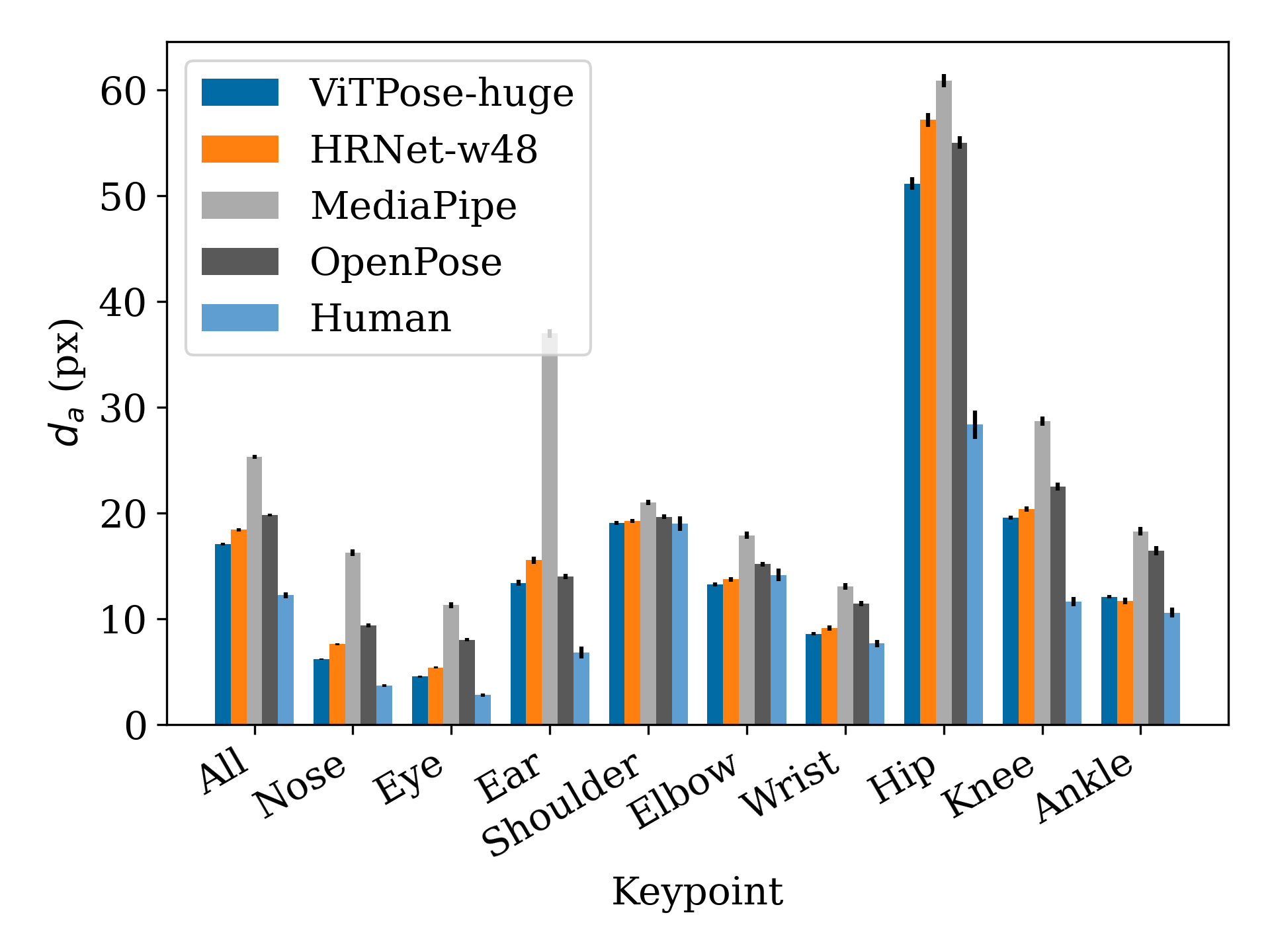}
        \caption{Mean $d_a$ of the different generic frameworks, human for comparison.}
        \label{fig:modelerror}
    \end{subfigure}
    \begin{subfigure}[t]{0.49\linewidth}
        \includegraphics[width = \linewidth]{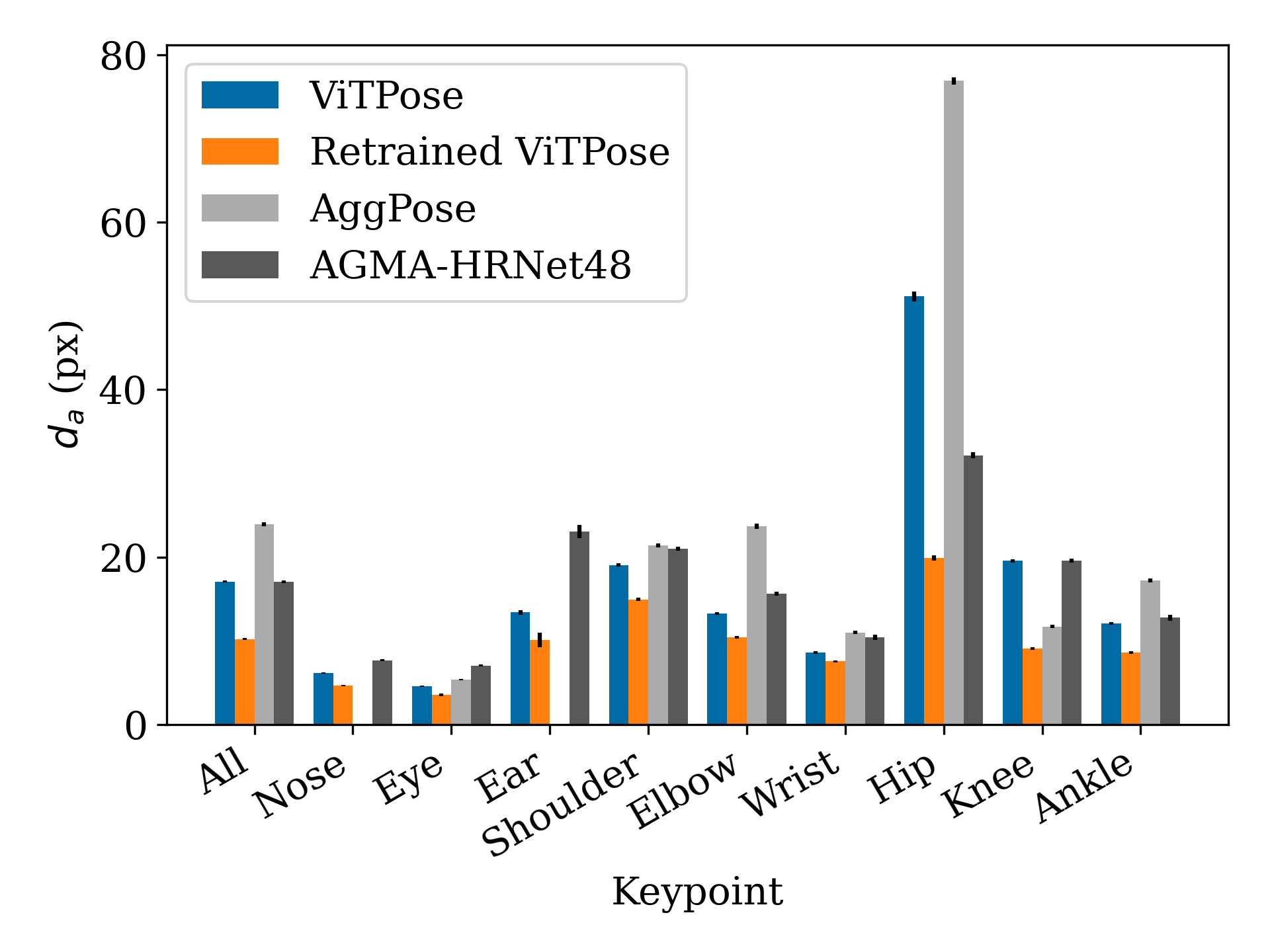}
        \caption{Mean $d_a$ for the infant pose estimators. Generic ViTPose is also shown for comparison to non-retrained results. Note that AggPose does not estimate the positions of nose and ears. }
        \label{fig:retrained_modelerror}
    \end{subfigure}
    \caption{Difference to annotation $d_a$ in pixels for different subjects, evaluated on our dataset and grouped by key point. Error bars represent confidence intervals of mean (95\%).}
\end{figure*}

All four generic pose estimation models were used to estimate keypoints in the whole dataset.
The difference between the predictions and the ground truth is shown in Figure \ref{fig:modelerror}.
The human labeling difference as presented in Section \ref{sec:humanvariation} is added for comparison.
Table \ref{tab:PCK} shows the corresponding PCK values.

The qualitative result is very close to the difference between human annotators, with the hips having the highest error.
But in contrast to the human annotations, the shoulder error is relatively lower, compared to the hip error.
Because of the size of the dataset, the differences between the pose estimation models are all statistically significant with $p<0.001$  (paired sample t-test on all keypoints).

The current state of the art model on the COCO dataset, ViTPose-huge, is also the best performing model on our dataset.
However, the relative distance to the second best, HRNet-w48, is only 2.4\% for PCK@0.1.
For high accuracy requirements, e.g., PCK@0.05, the relative distance is only 0.6\%, but the difference is still statistically significant with $p=0.0312$ (Pearson's chi-squared test).
OpenPose, although being an older model, still achieved better results than MediaPipe, which did never achieve a mean error of less than 10 pixels, not even for the clearly defined keypoints like eye and nose.

\subsection{Infant pose estimators}

We evaluated the retrained pose estimation models in the same way we evaluated the generic ones.
Figure \ref{fig:retrained_modelerror} shows the position estimation errors for the different models with the non-retrained ViTPose network for comparison.
The corresponding overall PCK results are displayed in Table \ref{tab:PCK}.

\begin{table}[ht]
    \centering
    \caption{PCK values for pose estimation with the different models for the complete dataset. Upper section: generic models, lower section: retrained models}
    \label{tab:PCK}
    \renewcommand*{\arraystretch}{1.2}
    \begin{tabular}{llll}
        Model           & PCK@0.1       & PCK@0.075         & PCK@0.05          \\
        \hline
        ViTPose-huge    & \textbf{84.6} & \textbf{75.49}    & \textbf{59.50}    \\
        HRNet-w48       & 82.6          & 73.92             & 59.12             \\
        MediaPipe       & 70.97         & 58.91             & 39.99             \\
        OpenPose        & 79.48         & 70.15             & 53.74             \\
        \hline
        \hline
        Retrained ViTPose   & \textbf{93.89}    & \textbf{89.82}   & \textbf{79.64}    \\
        AggPose             & 75.60             & 67.02            & 52.43             \\
        AGMA-HRNet48        & 84.32             & 75.12            & 59.41             \\
    \end{tabular}
\end{table}

Retraining significantly improved the ViTPose model, with the PCK increasing by 20 percentage points in PCK@0.05 ($p<0.001$, Pearson's chi-squared test). 
The mean difference to annotation for the hips is decreased by 61\%.
All other keypoints also see significant improvements (all $p<0.001$, individual paired sample t-tests on keypoints).

The other two infant pose estimators Aggpose \cite{Cao_AggPoseDeepAggregation_2022} and AGMA-HRNet48 \cite{Soualmi_3Dposeestimation_2023} exhibit different behaviour.
Both perform significantly worse than our retrained network ($p<0.001$, paired sample t-test on all keypoints), with AggPose even performing worse than the non-retrained ViTPose ($p<0.001$, paired sample t-test on all keypoints).
The difference between ViTPose and AGMA-HRNet48 is not statistically significant ($p=0.66$ for $d_a$, paired sample t-test on all keypoints, and $p=0.59$ for PCK@0.05, Pearson's chi-squared test).
Regarding the hips, AGMA-HRNet48 improves the accuracy over generic pose estimation, while AggPose is significantly worse (both $p<0.001$, paired sample t-test on all keypoints).

\subsection{Influence of the view angle}
\label{sec:anglecomparison}

\begin{figure*}[ht]
    \centering
    \begin{subfigure}[t]{0.49\linewidth}
		\includegraphics[width=\textwidth]{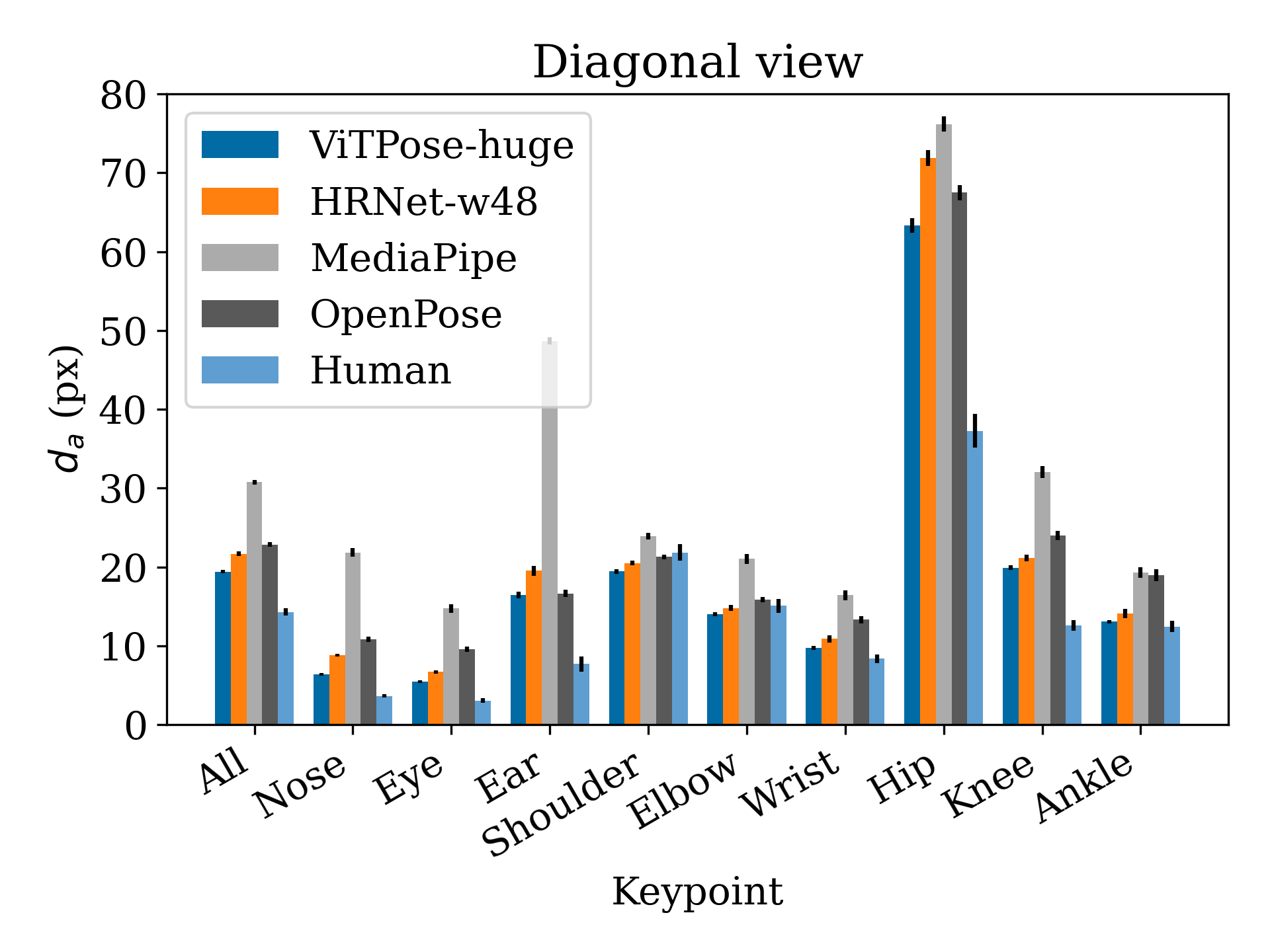}
	\end{subfigure}
	\begin{subfigure}[t]{0.49\linewidth}
		\includegraphics[width=\textwidth]{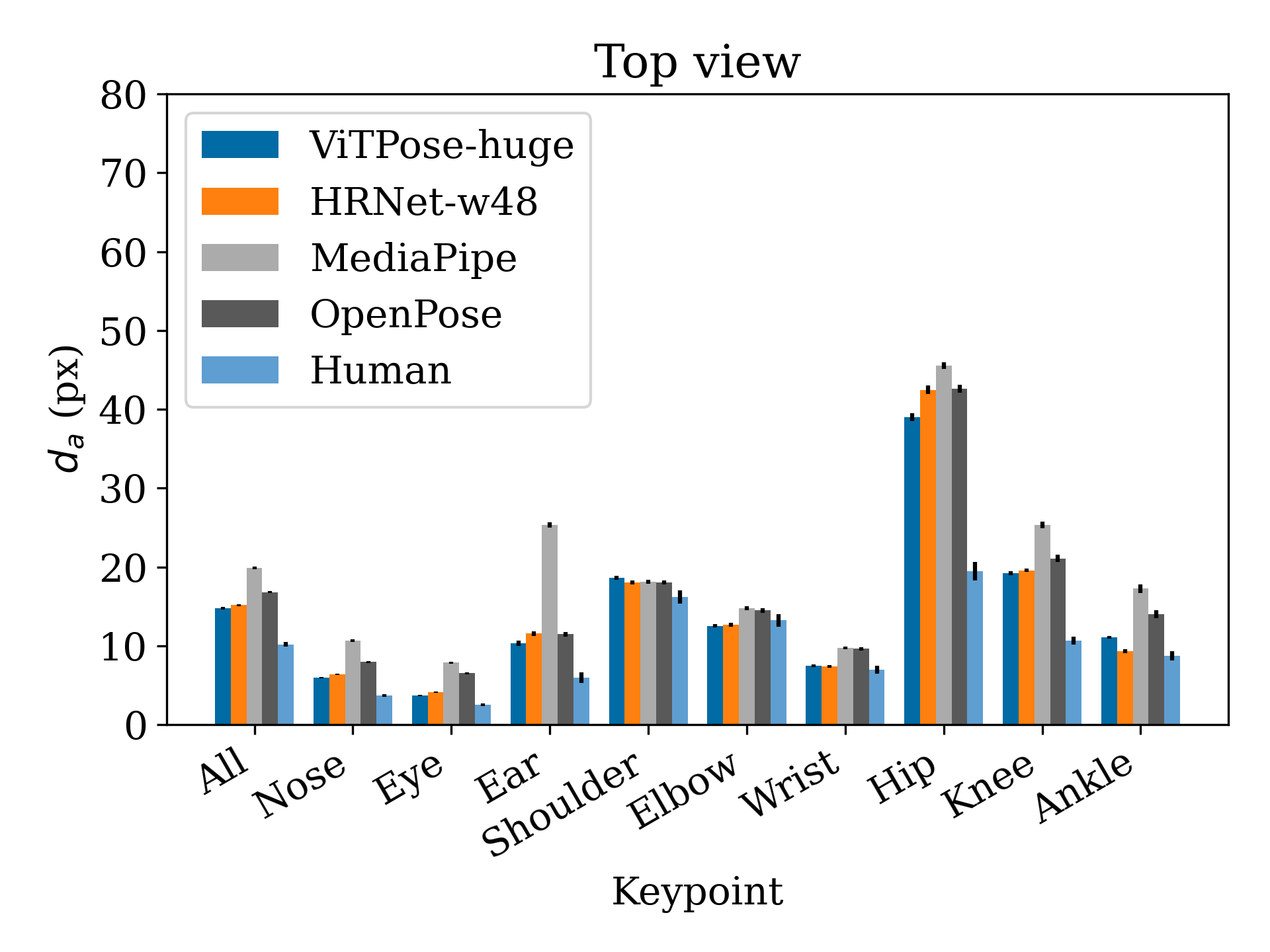}
	\end{subfigure}
    \caption{Pose estimation errors for the generic pose estimation models, split by viewing angle. Error bars represent confidence intervals of mean (95\%).}
    \label{fig:modelerror_anglesplit}
\end{figure*}

Figure \ref{fig:modelerror_anglesplit} shows the pose estimation error as Figure \ref{fig:modelerror}, but split into top and diagonal view.

The errors in case of the diagonal view are higher than in the case of the top view. All top/diagonal differences are statistically significant with $p<0.001$ (paired sample t-test on all keypoints).
The most pronounced difference is between the hip position estimation errors.
In the top view, the difference is not only lower than in the diagonal view, but also lower as compared to the errors of the other keypoints.
This is consistent with the results for the difference between human annotators shown in Figure \ref{fig:humanerror}.

MediaPipe does not show any benefit from being designed for fitness applications.
In contrary, its performance decreases the most for the diagonal view.

\subsection{Retraining for individual view angles}

\begin{figure*}[ht]
    \centering
    \begin{subfigure}[t]{0.49\linewidth}
		\includegraphics[width = \linewidth]{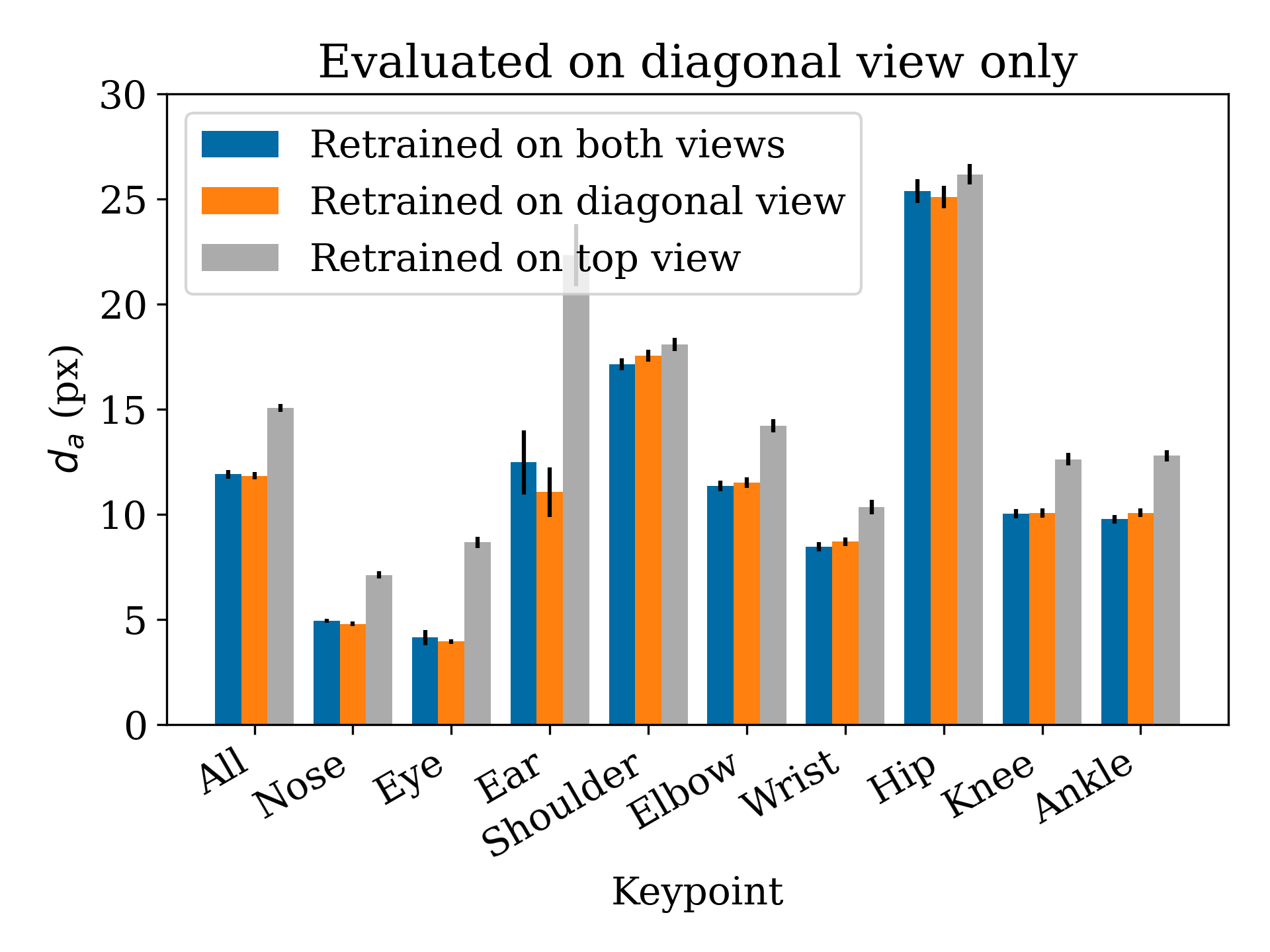}
        \label{fig:retrain_error_split_c1}
	\end{subfigure}
	\begin{subfigure}[t]{0.49\linewidth}
		\includegraphics[width = \linewidth]{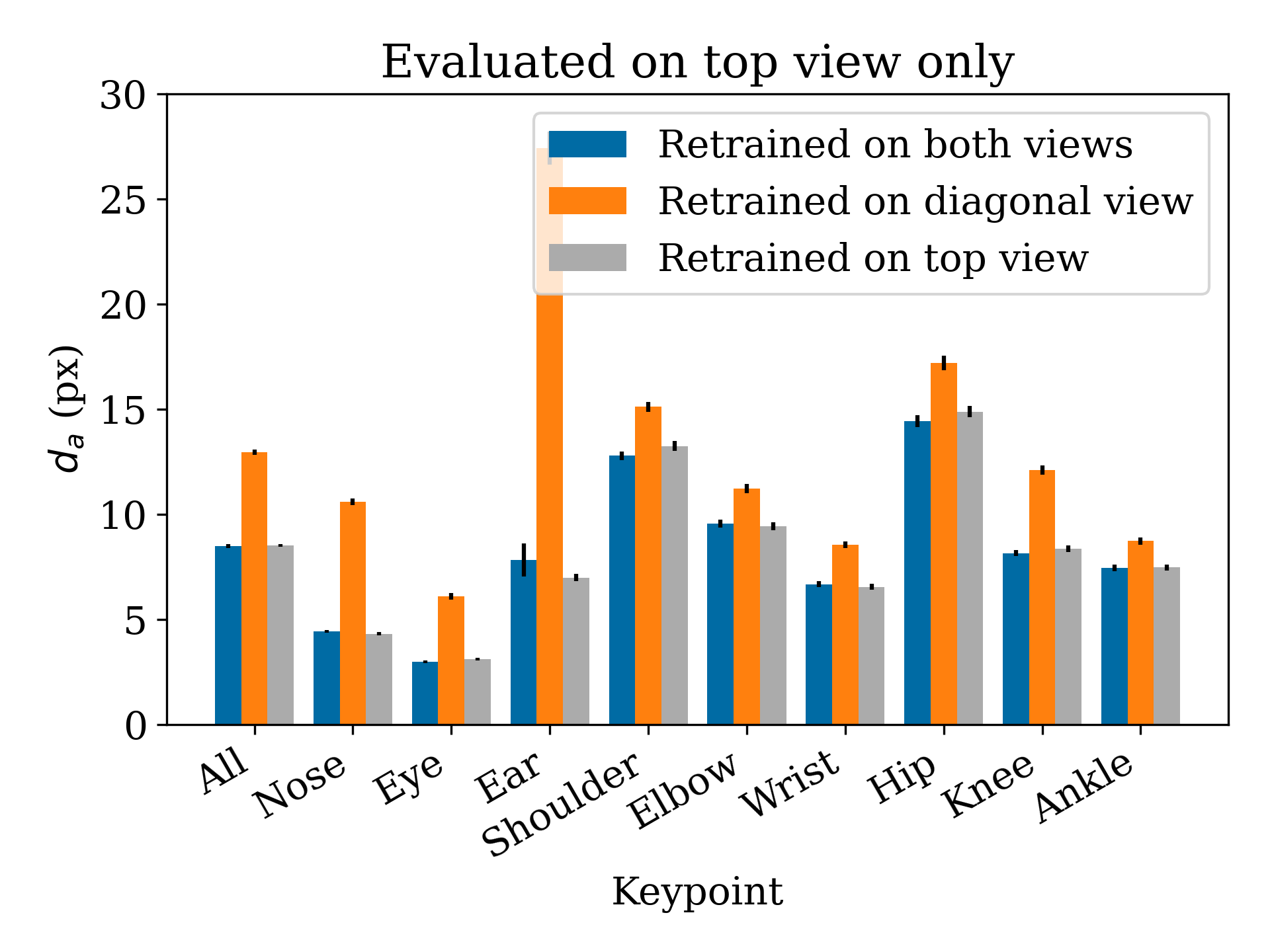}
        \label{fig:retrain_error_split_c2}
	\end{subfigure}
    \caption{Mean difference from annotation ($d_a$) for retrained ViTPose models separately evaluated on diagonal or top view images. Different models have been trained on all, only diagonal view or only top view images, respectively. Error bars represent confidence intervals of mean (95\%).}
    \label{fig:retrained_angleerror}
\end{figure*}

As was shown in Section \ref{sec:anglecomparison}, there is a significant difference between the diagonal and top views.
We retrained ViTPose models only on the individual viewing angles, to see if specializing on them helps to improve the performance.
The results are presented in Figure \ref{fig:retrained_angleerror}.

There is no significant difference between the model that was trained on both views versus the ones trained on individual views ($p=0.25$ for diagonal, $p=0.54$ for top view, both paired sample t-tests on all keypoints), if evaluated against the respective views that were used in training.
The results for the respective views that were not used in training, however, are significantly worse ($p<0.001$, paired sample t-test on all keypoints).
This means there is no gain in specializing models to certain viewing angles in our case.
However, training on a view that is different from the one used in inference leads to worse performance.

Regarding the general difference in performance between the viewing angles (without specialized models), we again observe worse results in the diagonal view.
Even with retraining, it is not possible to achieve the same accuracy of the diagonal view as of the top view.

\subsection{Inference speed}
Since the focus of our study is not on real-time  pose estimation for real time applications, we did not optimize the pose estimators (e.g., architectures and/or meta parameters) for inference speed.

All models can be executed on any recent GPU with at least 4GB of memory (for model complexity please see the respective original publications). 
Table \ref{tab:inference} lists the observed inference speed on a NVIDIA GTX 2080 Ti.
Those speeds were obtained by using the demonstration scripts of the respective models and adapting them to process our dataset.
The results show that all models have acceptable and reasonable inference speed with the ViTPose being the slowest (but the most accurate), and OpenPose the fastest model.

\begin{table}[ht]
    \centering
    \small
    \caption{Inference speed in frames per second (fps), on a NVIDIA GTX 2080 Ti, for each included model. This is the average speed over processing the 4500 frames of the dataset.}
    \label{tab:inference}
    \renewcommand*{\arraystretch}{1.2}
    \begin{tabular}{l|l|l|l|l|l|l|l}
        Model           &ViTPose-huge   &HRNet-w48  &MediaPipe      &OpenPose   &Retrained ViTPose  &AggPose    &AGMA-HRNet48 \\
        \hline  
        Inference speed &1\,fps          &1.2\,fps    &3\,fps (CPU)    &11.5\,fps   &1.1\,fps            &2.7\,fps    &3\,fps        \\
    \end{tabular}
\end{table}

\section{Discussion}
\label{sec:discussion}

There is a wealth of tools for infant movement classification available that are all trained and tested on silo-datasets mostly using one camera view, and thus not directly comparable \cite{Marschik_Openvideodata_2023}. We therefore used a newly designed multi-view dataset that allowed for a direct comparison of the available models. 

\paragraph{Generic pose estimators}
The state of the art models ViTPose and HRNet show clear improvements compared to the older model OpenPose.
The gain in PCK@0.05 of 5.76 percentage points is small compared to the gain later achieved by retraining, but still statistically significant ($p<0.001$, Pearson's chi-squared test).

In addition to the position accuracy, the reliability of keypoint detection also increases with newer models (see Figure S1). A discussion is given in the Appendix.

\paragraph{Infant specific pose estimators}
The comparison with the other infant pose estimators showed that the generalization capabilities of neural networks are overestimated in this field of study.
Our results show that the specialized infant pose estimators do not perform substantially better on our infant dataset than the generic ViTPose model.
AGMA-HRNet48 showed no significant overall difference to generic ViTPose, but improved on the hip keypoint. AggPose however, performed worse for all keypoints.

While the AGMA-HRNet48 does improve pose estimation accuracy of the hip, other keypoints get worse, so there is no statistically significant difference between AGMA-HRNet48 and generic ViTPose ($p=0.66$, paired sample t-test on all keypoints).
It has to be noted that the training data of this model and our training data were both derived from a GMA setting.
They differ in the amount of visual clutter in the image (compare with Figure 3 in \cite{Soualmi_3Dposeestimation_2023}), but otherwise closely resemble each other.
The infant specific retraining improved the HRNet48 by 0.29 and 1.72 percentage points in PCK@0.05 and PCK@0.1 on our dataset, respectively, putting AGMA-HRNet48 on the same performance level as the generic ViTPose.
Note, that the AGMA-HRNet48 model is not the best performing one from its original paper \cite{Soualmi_3Dposeestimation_2023}.
Their DarkPose32 model scored 2.23 and 0.55 percentage points higher than HRNet48 at PCK@0.05 and PCK@0.1, respectively.
Since the non-retrained DarkPose32 model performed worse than the non-retrained HRNet48 in \cite{Soualmi_3Dposeestimation_2023}, we do not expect it to yield a significant improvement over the generic pose estimation on our data either.

AggPose also uses Vision Transformers, like ViTPose, but is trained on a substantially larger dataset than our retrained model (20748 images, \cite{Cao_AggPoseDeepAggregation_2022}), still it performs even worse than generic ViTPose, which suggests it is also overfit and cannot sufficiently generalize to our (and potentially to other) infant dataset although the underlying dataset is also from a GMA related setting.
AggPose has been compared to other generic pose estimators by Yin et al. (see \cite{Yin_selfsupervisedspatiotemporalattention_2024}, Tables 1, 2 and 3), where it also performs worse than the generic pose estimators on different infant pose datasets.

To date, there are no publicly available infant datasets suitable for GMA (mostly due to patient data privacy issues; see \cite{Marschik_Openvideodata_2023}), so we could not evaluate our model on data from a different setup.
This is probably further contributing to the general overfitting situation we identified, because there is no training set available that covers multiple GMA related setups. See the Appendix for a comment on a preliminary analysis with the SyRIP dataset \cite{Huang_InvariantRepresentationLearning_2021}.

\paragraph{Influence of viewing angle}
There is a significant influence of the viewing angle on pose estimation accuracy.
Every performance, human, generic and retrained models, is worse on the diagonal view than on the top view.
We see two main reasons for this.

One reason for the reduced performance is a lack of training data from the corresponding view angle.
For everyday situations like those in the COCO dataset, the viewing angle of the diagonal camera corresponds to placing the camera very close to the floor.
This is uncommon.
We included the MediaPipe framework in our evaluation, because it is marketed towards fitness applications and for the use on mobile devices, a common application in GMA \cite{Marschik_MobileSolutionsClinical_2023}.
This suggested to us that it might be trained on images with uncommon view angles (e.g., mobile phone put on the ground to observe the fitness-workout).
The training dataset even contained 25000 frames from fitness exercises \cite{Bazarevsky_BlazePoseOndeviceRealtime_2020}.
However, the model performed worse compared to all others, with no beneficial effect on the diagonal angle.
It has to be said though, that MediaPipe is a small model (with respect to the number of parameters) designed for inference on mobile devices and as such not as potent as the other models.

Another reason for the better results of the top view angle is occlusion of features.
The anatomy of the baby makes it easy to align the arms and legs with the sight line axis of the camera when lifting them, blocking the view on body keypoints like knee, wrist, shoulder or hip (see the second example from the left in Figure \ref{fig:setup_overview}b,c).
It might be preferable for computer vision to use a diagonal angle from the head end of the bed, instead of the foot end, to make the extremities lift into an orthogonal direction to the camera sight line.

\paragraph{Retraining to individual viewing angles}
Even when retraining the model to the specific view angle, we could not achieve the same performance on the diagonal view as on the top view (see Figure \ref{fig:retrained_angleerror}).
We suppose this is because the human variation in the training data was still higher in the diagonal view (see Figure \ref{fig:humanerror}), because of occlusions.
An interesting follow-up would be to only indirectly label both top and diagonal view via reprojection from the side views (which might be hard to do, as they are only the supporting views for occlusion resistance) and see if the significant difference between the view angles persists.

\section{Conclusion}
\label{sec:conclusion}

We compared four different generic pose estimation frameworks on our dataset of 4500 frames recorded with a multi-view camera setup designed for automated GMA.
The current best performing model on COCO test-dev, ViTPose, also achieved the best results on our infant dataset.
Retraining the best performing network on our dataset, expectedly, increased the performance, especially on the keypoint that is most difficult to detect, the hips. 
In comparison with the other infant pose estimation frameworks, we achieved 20 percentage points better PCK@0.05 than the second best model.
The other models performed below or on the level of the best generic pose estimator.
This suggests that, if possible, one should always retrain on the specific dataset to achieve the best pose estimation accuracy. If retraining is not possible, a state-of-the-art generic pose estimator should be used, unless the dataset is very similar to the training data of a specialized infant pose estimator.

We also compared the pose estimation error between different viewing angles.
There is a significant difference in accuracy between the diagonal and top views in both human annotation and neural network pose estimation.
Therefore, the results strongly suggest that the top view should be preferred in any setup for \textit{automated} GMA with only one camera.
Since the different viewing angles lead to different accuracy, we tested if specialized pose estimators trained on particular views would improve the accuracy, however, no significant effect was found.

In summary, our study suggests that pose estimation accuracy would benefit from the utility of the top view and the pose estimator being trained on the specific dataset. However, a generic pose estimator should be preferred over a specialized pose estimator trained on a different dataset, if retraining is not feasible. Whether potentially also movement classification accuracy would be improved with more accurate pose estimation remains a question, which needs to be addressed in future work. While the standard GMA method uses a diagonal view for assessment, clinical setups aiming to generate data for AI and computer vision approaches should in future consider top-down viewing angles for recording new data.

\section*{Acknowledgements}
We would like to thank the participating families in the SEE-study (Systemic Ethology and Developmental Science) and the additional annotator Felicia Seita; this study was conducted within the SFB 1528, project C03, DFG; clinical aspects are related to VW-IDENTIFIED and the DFG P456967546. We were supported by the FWF KLI811. We would like to thank all SEE (Systemic Ethology and Developmental Science, Marschik Lab) members who have contributed to this study.

\section*{Author contributions statement}

L.J. P.M. and T.K developed the idea and research questions. 
L.J. performed the analysis, on data annotated by S.F., and wrote the main manuscript text. 
T.K. supervised the work and edited the manuscript from the first draft onward. 
D.Z., L.P. and P.M. wrote the sections about the clinical aspects and GMA. 
S.F., D.Z. S.B. F.W. and P.M. edited the manuscript. 
All authors read and approved the final manuscript version.

\section*{Data availability statement}
The authors do not have permission to share the infant video recording data. Derived data, like the annotations and pose estimation data are available from the corresponding author upon reasonable request. The five retrained ViTPose models are published on Zenodo (\url{https://doi.org/10.5281/zenodo.14833182}).

\section*{Additional information}

\textbf{Competing interests:} The authors declare no competing interests.

\renewcommand{\thefigure}{S\arabic{figure}}
\setcounter{figure}{0}
\appendix
\section{Appendix}

\subsection{Keypoint detection rates}
\label{sup:detectionrates}

This section contains an additional analysis of the models own estimation of reliability.
All models also score the predictions with a certainty value $c$ between 0 and 1, e.g., how confident the model is in its prediction.
This value in itself is not normalized between the models.
To enable comparison, we do not look at these certainty values, but at the level of missing keypoints.
To obtain it, we thresholded the certainty values, for thresholds $t$ between 0 and 1, and then calculated the respective ratio of missing predictions $m$ as
\begin{equation}
    m = \frac{1}{N} \sum \limits_{i=1}^N H(c-t). \nonumber
\end{equation}
There, $H$ is the Heaviside function and $N$ the total number of annotated keypoints in the dataset.
This effectively normalizes the score to the percentage of points deemed correct by the model itself.
We then plotted $d_a$ against this dependent value $m$ instead of the model output $c$.

\begin{figure*}[ht!]
    \centering
    \begin{subfigure}[t]{0.49\linewidth}
		\includegraphics[width = \linewidth]{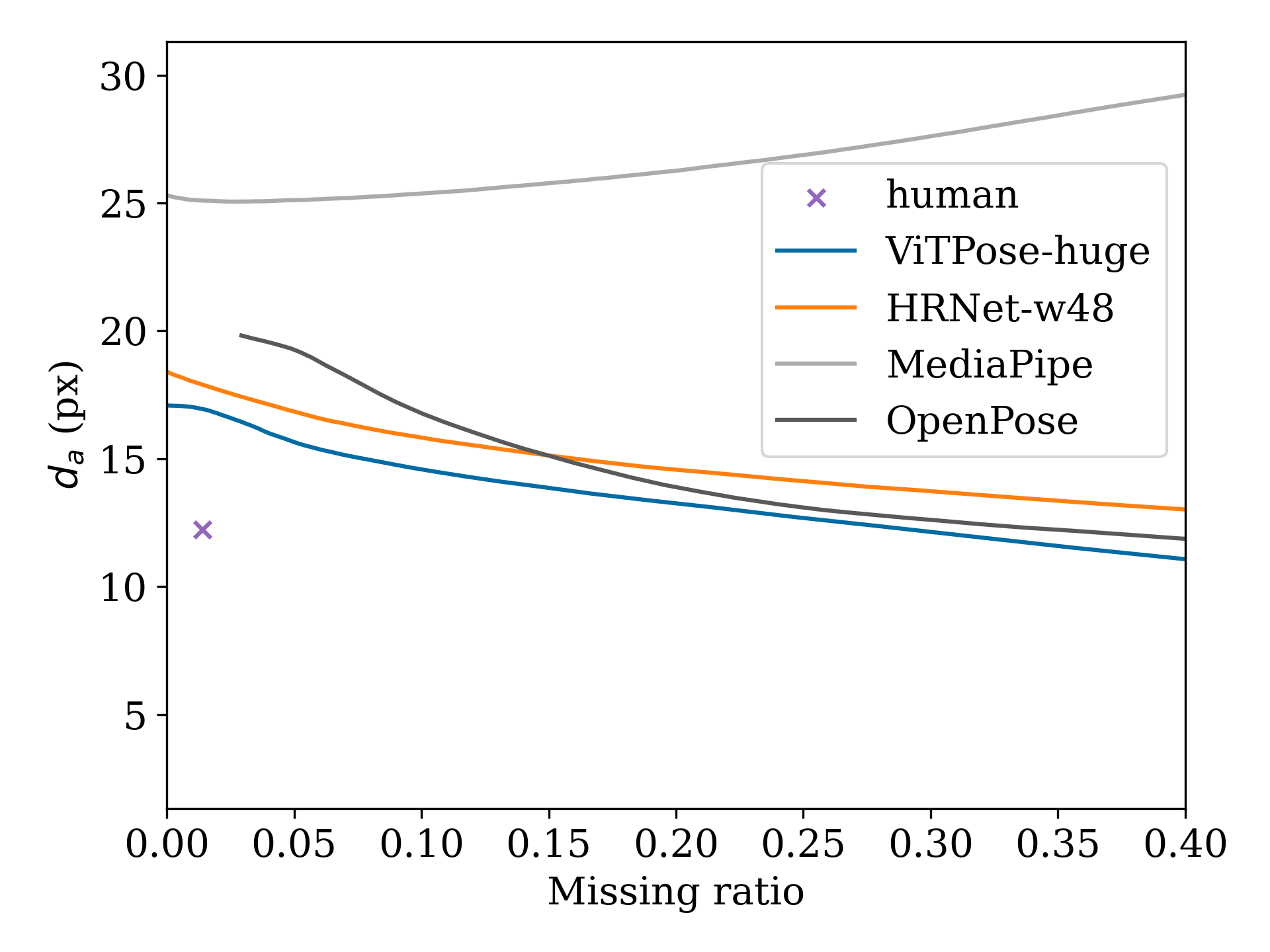}
        \caption{Overall error}
        \label{fig:scorecurve}
	\end{subfigure}
	\begin{subfigure}[t]{0.49\linewidth}
		\includegraphics[width = \linewidth]{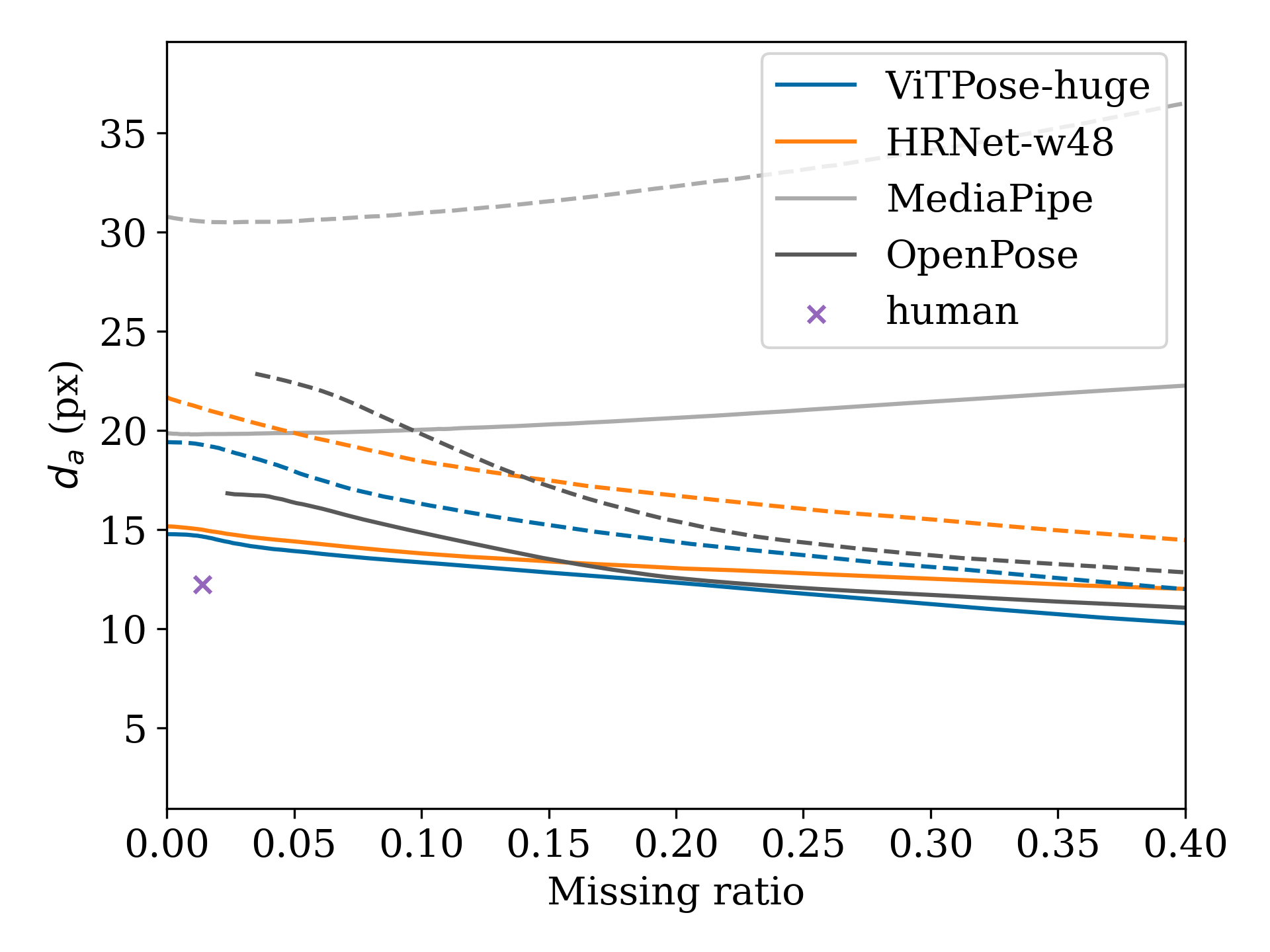}
        \caption{Split by viewing angle}
        \label{fig:scorecurve_split}
	\end{subfigure}
    \caption{Mean difference to annotation $d_a$ vs. the ratio of missing detections, based on the keypoint prediction confidence scores. Panel a): overall, panel b): split by viewing angle, solid lines correspond to the top view, dashed lines to the diagonal view.}
    \label{fig:sc}
\end{figure*}

The keypoint detection rates for all models are displayed in Figure \ref{fig:scorecurve}.
The result for the human annotation variance (see Section 3.1) was also added for comparison.

Except for MediaPipe, the error decreases when filtering out uncertain points.
The increase for MediaPipe comes from the model assigning relatively high scores to the ears, which are among its worst detected points.
ViTPose-huge has the lowest error across all missing ratios, MediaPipe the highest.
OpenPose achieves a lower error than HRNet for missing ratios over 15\%, but this is of limited use, as ViTPose is still better and 15\% of points not being detected is insufficient for the applications of automated GMA.
Moreover, OpenPose does not even detect all points when thresholding with a score of 0, resulting in the missing ratio never dropping below 2.93\%.
ViTPose and HRNet, however, produce predictions for every possible point when thresholding at 0.
Below the missing ratio for humans (1.41\%, because of the ears), the error of ViTPose saturates, while the error of HRNet increases.
This is because ViTPose is assigning low scores to the ear keypoints that are not visible (e.g., due to head turned to the side) and therefore don't affect the error calculation.

Figure \ref{fig:scorecurve_split} shows the mean difference to annotation in dependence of the missing ratio, like Figure \ref{fig:scorecurve}, but split by perspective.
The performance for the diagonal view is always worse than for the top view.
Moreover, while the error for the diagonal view drops faster than for the top view when filtering out uncertain points, they don't converge to the same level even when filtering out more than 20\% of all detected points.
We again observe OpenPose not being able to provide estimates for every point, with the diagonal angle having more non-detected keypoints as compared to the top view.

In summary, filtering out points with low certainty scores yields better results in terms of pose estimation error ($d_a$) for all models but MediaPipe.
However, this is of limited value for practical use, since, for most applications, missing values would have to be filled in by interpolation, median filtering, or other techniques (e.g., Kalman filter), to be used for motion analysis.
Still, the fact that the least certain points for ViTPose are the ones humans could not annotate, manifesting in stagnating $d_a$ for low missing ratios, shows the certainty score aligns with actual visibility constraints for the state-of-the-art model.

\subsection{SyRIP}
\label{sup:syrip}

The SyRIP dataset$^{40}$ was considered, and some preliminary analyses were performed, however, it became obvious that the setting of this dataset is too different from GMA setting, with many more different body positions and much older infants than in our dataset.
As in case of the specialized infant pose estimators evaluated on our dataset, our model could not compete with the generic ViTPose on SyRIP.
Retraining decreased the PCK@0.05 from 51.53\% to 27.65\% (73.66\% to 46.97\% for PCK@0.1), further strengthening our point that specialized infant pose estimators (in this case our own) are overfit and do not generalize well to other datasets.

\end{document}